\def\paperID{0000}
\def\confName{CVPR}
\def\confYear{2024}

\newif\ifreview 
\newif\ifarxiv \newcommand{\arxiv}{\arxivtrue}
\newif\ifcamera 
\newif\ifrebuttal 

\arxiv

\pdfoutput=1
\documentclass[10pt,twocolumn,letterpaper]
{article}
\ifreview \usepackage[review]{cvpr} \fi
\ifarxiv \usepackage[pagenumbers]{cvpr} \fi
\ifrebuttal \usepackage[rebuttal]{cvpr} \fi
\ifcamera \usepackage{cvpr} \fi

\usepackage{graphicx}	
\usepackage{amsmath}	
\usepackage{amssymb}	
\usepackage{booktabs}
\usepackage{times}
\usepackage{microtype}
\usepackage{epsfig}
\usepackage[table,xcdraw,dvipsnames]{xcolor}
\usepackage{caption}
\usepackage{float}
\usepackage{placeins}
\usepackage{color, colortbl}
\usepackage{stfloats}
\usepackage{enumitem}
\usepackage{tabularx}
\usepackage{xstring}
\usepackage{multirow}
\usepackage{xspace}
\usepackage{url}
\usepackage{subcaption}
\usepackage{xcolor}
\usepackage{soul}
\usepackage[hang,flushmargin]{footmisc}

\ifcamera \usepackage[accsupp]{axessibility} \fi

% ...

\ifarxiv  \fi

\newcommand{\R}[1]{{%
    \textbf{%
        \ifstrequal{#1}{1}{\textcolor{red}{R#1}}{%
        \ifstrequal{#1}{2}{\textcolor{blue}{R#1}}{%
        \ifstrequal{#1}{3}{\textcolor{magenta}{R#1}}{%
        \ifstrequal{#1}{4}{\textcolor{teal}{R#1}}{%
                           \textcolor{cyan}{R#1}%
        }}}}%
    }%
}}

\usepackage[dvipsnames]{xcolor}
\usepackage{xr-hyper}

\makeatletter
\newcommand*{\addFileDependency}[1]{
  \typeout{(#1)}
  \@addtofilelist{#1}
  \IfFileExists{#1}{}{\typeout{No file #1.}}
}

\makeatother
\newcommand*{\myexternaldocument}[1]{
    \externaldocument{#1}
    \addFileDependency{#1.tex}
    \addFileDependency{#1.aux}
}

\definecolor{cvprblue}{rgb}{0.21,0.49,0.74}
\usepackage[pagebackref,breaklinks,colorlinks,citecolor=cvprblue]{hyperref}
\usepackage[capitalize]{cleveref}
\crefname{section}{Sec.}{Secs.}
\crefname{table}{Table}{Tables}
\crefname{figure}{Fig.}{Figs.}

\ifarxiv \crefname{appendix}{App.}{Apps.}
\else \crefname{appendix}{Suppl.}{Suppls.} \fi

\usepackage{adjustbox}

\unless\ifarxiv \myexternaldocument{_supplementary} \fi

\begin{document}
%% TITLE
\title{LAESI: Leaf Area Estimation with Synthetic Imagery}

\author{\hspace{1.25cm}Jacek Ka\l{}u\.zny\\
\hspace{1.25cm}AMU\\
\hspace{1.25cm}{\tt\small jacek.kaluzny@amu.edu.pl}
\and
\hspace{0.5cm}Yannik Schreckenberg\\
\hspace{0.5cm}TUM\\
\hspace{0.5cm}{\tt\small yannik.schreckenberg@tum.de}
\and
\hspace{1.0cm}Karol Cyganik\\
\hspace{1.0cm}AMU\\
\hspace{1.0cm}{\tt\small karol.cyganik@amu.edu.pl}
\and
\hspace{0.75cm}Peter Annigh\"ofer\\
\hspace{0.75cm}TUM\\
\hspace{0.75cm}{\tt\small peter.annighoefer@tum.de}
\and
\hspace{2.0cm}S\"oren Pirk\\
\hspace{2.0cm}GreenMatterAI / CAU\\
\hspace{2.0cm}{\tt\small soeren.pirk@greenmatter.ai}
\and
\hspace{-0.4cm}Dominik L. Michels\\
\hspace{-0.4cm}GreenMatterAI / KAUST / TU Darmstadt\\
\hspace{-0.4cm}{\tt\small dominik.michels@greenmatter.ai}
\and
\hspace{0.3cm}Mikolaj Cieslak\\
\hspace{0.3cm}GreenMatterAI\\
\hspace{0.3cm}{\tt\small mikolaj.cieslak@greenmatter.ai}
\and
\hspace{0.75cm}Farhah Assaad-Gerbert\\
\hspace{0.75cm}TUM\\
\hspace{0.75cm}{\tt\small farhah.assaad@tum.de}
\and
\hspace{2.75cm}Bedrich Benes\\
\hspace{2.75cm}Purdue University\\
\hspace{2.75cm}{\tt\small bbenes@purdue.edu}
\and
\hspace{1.0cm}Wojciech Pa\l{}ubicki\\
\hspace{1.0cm}GreenMatterAI / AMU\\
\hspace{1.0cm}{\tt\small wojciech.palubicki@greenmatter.ai}
}

\maketitle

\begin{abstract}
% Abstract goes here.
%We introduce LAESI the Synthetic Leaf Dataset, a collection of 100,000 synthetic leaf images on millimeter paper, each with semantic masks and surface area labels. 
We introduce LAESI, a Synthetic Leaf Dataset of 100,000 synthetic leaf images on millimeter paper, each with semantic masks and surface area labels. This dataset provides a resource for leaf morphology analysis primarily aimed at beech and oak leaves. We evaluate the applicability of the dataset by training machine learning models for leaf surface area prediction and semantic segmentation, using real images for validation. Our validation shows that these models can be trained to predict leaf surface area with a relative error not greater than an average human annotator. LAESI also provides an efficient framework based on 3D procedural models and generative AI for the large-scale, controllable generation of data with potential further applications in agriculture and biology. We evaluate the inclusion of generative AI in our procedural data generation pipeline and show how data filtering based on annotation consistency results in datasets which allow training the highest performing vision models. %The dataset and the code of the generative framework are publicly available, supporting further research and application in this field.
\end{abstract}
\section{Introduction}
\label{sec:intro}
In recent years, the integration of machine vision algorithms in agriculture has been instrumental in enhancing productivity and sustainability, and novel machine learning-based methods have allowed for solving problems that were impossible ten years ago. One limitation of the machine learning algorithms is their reliance on accurate and extensive training data. This is because acquiring sufficiently annotated real-world data, particularly for tasks like leaf analysis, is often costly and time-consuming~\cite{ward2018deep}.

We present LAESI, a Synthetic Leaf Dataset, and two procedural models for its generation. The first is a procedural model for millimeter paper, and the other is for leaf shape generation. Our method employs ControlNet~\cite{zhang2023controlnet} to improve the visual realism of renderings similarly to \citeauthor{Anagnostopoulou_2023_CVPR} \cite{Anagnostopoulou_2023_CVPR}. Using computationally efficient procedural models paired with generative AI models allows for a fully automatic, controllable, and large-scale generation of synthetic data that is useful for training deep learning models for vision tasks.

The procedural model for millimeter paper provides a scalable background for each image, enabling consistent leaf surface area annotations for data points. The leaf model employs procedural generation, which simulates a wide variety of leaf types with realistic shapes, and the additional appearance model generates a wide variety of textures.

The LAESI dataset provides annotations consisting of semantic masks and surface area labels, which are useful for leaf morphology analysis. We demonstrate the utility of this dataset by training machine vision models on leaf surface area prediction and semantic segmentation. Furthermore, we compare vision models trained with different blends of synthetic data and real data against a baseline model trained on 1,7K real annotated images.

\begin{figure*}[htbp]
\centering
\includegraphics[width=\textwidth]{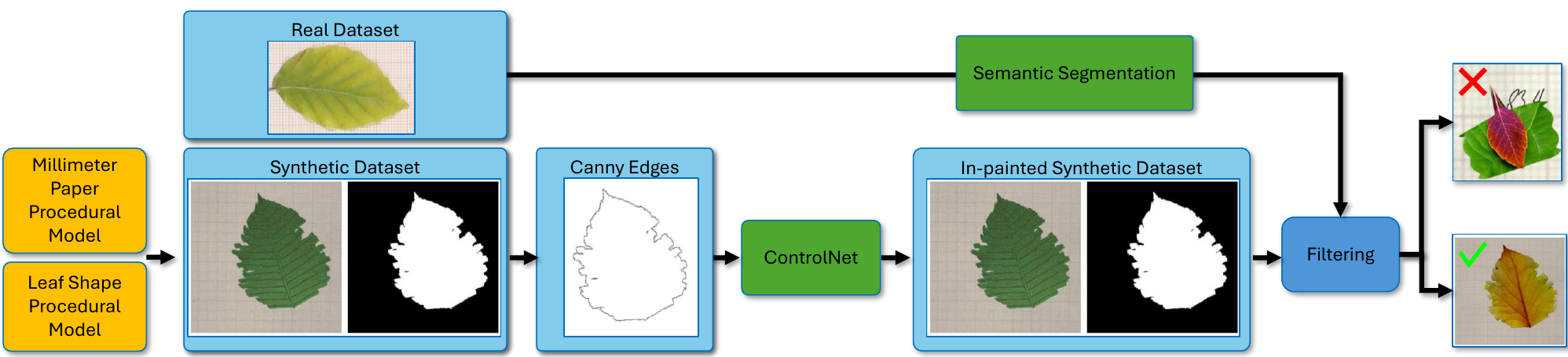}
\caption{A model of the LAESI pipeline: \textit{Procedural Generation of Millimeter Paper Background and Leaf Shape} generates diverse paper textures, grid alignments, and a range of leaf shapes, sizes, and textures. \textit{Rendering and Final Synthetic Dataset Composition} combines leaves with the background with realistic lighting and generates annotations such as semantic masks, surface area labels, and canny edges. \textit{Dataset Inpainting} utilizing the ControlNet-based pipeline for inpainting of Canny edges generates leaf images inside the masked regions of data points. \textit{Dataset Filtering} discards the leaf data points with inpainting results that reduce consistency with their annotations by using a semantic segmentation model.}
\label{fig:LAESI_pipeline}
\end{figure*}

Our pipeline for LAESI efficiently generates realistic and diverse synthetic leaf images through several stages (Fig.~\ref{fig:LAESI_pipeline}):
\textit{(1) Procedural Generation of Millimeter Paper Background:} generate various paper textures and grid alignments to ensure a consistent scale reference across the dataset;
\textit{(2) Leaf Shape and Texturing Procedural Model:} generates a wide range of leaf shapes, sizes, and textures to increase the dataset's variability;
\textit{(3) Semantic Mask and Surface Area Labeling:} Following leaf generation, semantic masks delineate leaf boundaries against the millimeter paper, paired with accurate surface area labels;
\textit{(4) Rendering and Final Image Composition:} the synthetic leaves are combined with the millimeter paper background, with an emphasis on realistic lighting, shadow effects, and overall image composition;
\textit{(5) Dataset Inpainting:} Each image in the dataset is processed using a ControlNet-based \cite{zhang2023controlnet}
pipeline for accurate inpainting of leaf masks using canny edges and text prompts as input;
\textit{(6) Semantic Segmentation-based Quality Control:} Inpainted synthetic images are semantically segmented into leaf and background and undergo a comparison with the procedurally generated ground truth masks to establish annotation consistency after inpainting.

The LAESI dataset comprises 100K data points. Each image features one synthetic leaf on millimeter paper and variable rendering parameters.

\section{Related Work}\label{sec:related}
Deep learning neural models have shown strong progress in many areas, but they require a large volume of high-quality data for effective training. While data are abundant, annotated data are expensive and difficult to obtain, especially in natural sciences, where the variance of a single biological species can be significantly high in both shape and appearance (texture). Various approaches have been developed to address this challenge, including semi-supervised, self-supervised learning, and synthetic data generation.

One notable approach in synthetic data generation is DatasetGAN \cite{zhang2021datasetgan}, which proposed a pipeline involving initial image generation by StyleGAN, followed by manual annotation of a few images for a specific task, and then training a small model to produce similar segmentation masks from StyleGAN features. This method allows the generation of a large number of labeled images with minimal manual effort. BigDatasetGAN \cite{zhang2021datasetgan} extended this concept using BigGAN for generating a diverse range of images, scaling it to the complexity of datasets like ImageNet.

In the realm of complex scene generation, \citeauthor{yang2022modeling} \cite{yang2022modeling} proposed a method for image generation based on specific layouts, compressing RGB images into patch tokens and utilizing a Transformer with Focal Attention. \citeauthor{sun2022shift} \cite{sun2022shift} introduced SHIFT, a synthetic driving dataset with variations in weather, time of day, and densities of vehicles and pedestrians, using domain adaptation for realistic simulations. \citeauthor{yan2021crossloc} \cite{yan2021crossloc} developed a visual localization system using a synthetic data generation tool that blends real and synthetic worlds, generating data with multiple annotations.

Similarly, \citeauthor{Anagnostopoulou_2023_CVPR} \cite{Anagnostopoulou_2023_CVPR} developed a Realistic Synthetic Mushroom Scenes Dataset, addressing the challenges in mushroom harvesting robotics. Close to our approach is the work of \citeauthor{ubbens18} \cite{ubbens18}, who developed a synthetic leaf model of rosette plants for counting focusing on describing the whole plant morphology. Our method presents a targeted approach for the generation of a large dataset of leaves for various tree species. Furthermore,   \citeauthor{zhang2023controlnet} \cite{zhang2023controlnet} demonstrated advancements in text-to-image diffusion models by adding conditional control.

Leaf appearance and modeling have been studied by computer graphics for decades. \citeauthor{chiba1996visual} \cite{chiba1996visual} proposed a method for leaf coloring and arrangement. \citeauthor{wang2004physically} used physics to simulate leaf growth in \cite{wang2004physically} and leaf venation patterns have been studied in \cite{runions2005modeling,hong2005interactive}. A general approach for leaf shape development considering experimental data from developmental biology was proposed in \cite{runions2017common}. We are not aware of any systematic approach to the generation of large leaf datasets for deep learning. %\mc{Only somewhat related, but, there is a large dataset of real images of leaf diseases: https://github.com/pratikkayal/PlantDoc-Dataset.}

Our work builds on existing research by investigating the integration of efficient, controllable 3D procedural models similar to ones used in \citeauthor{raistrick2023infinite} \cite{raistrick2023infinite} into a pipeline leveraging generative AI models to train deep learning models for specialized vision tasks where annotated real-world data are scarce or expensive to acquire.
\section{Method}
\label{sec:method}
LAESI allows for fully automatic large-scale synthetic data generation by leveraging simple but computationally efficient 3D procedural models for rendering. We implemented these procedural models with Unity. In this section, we discuss the individual components of the LAESI modeling and rendering pipeline.
\begin{figure}[htbp]
\centerline{\includegraphics[width=\linewidth]{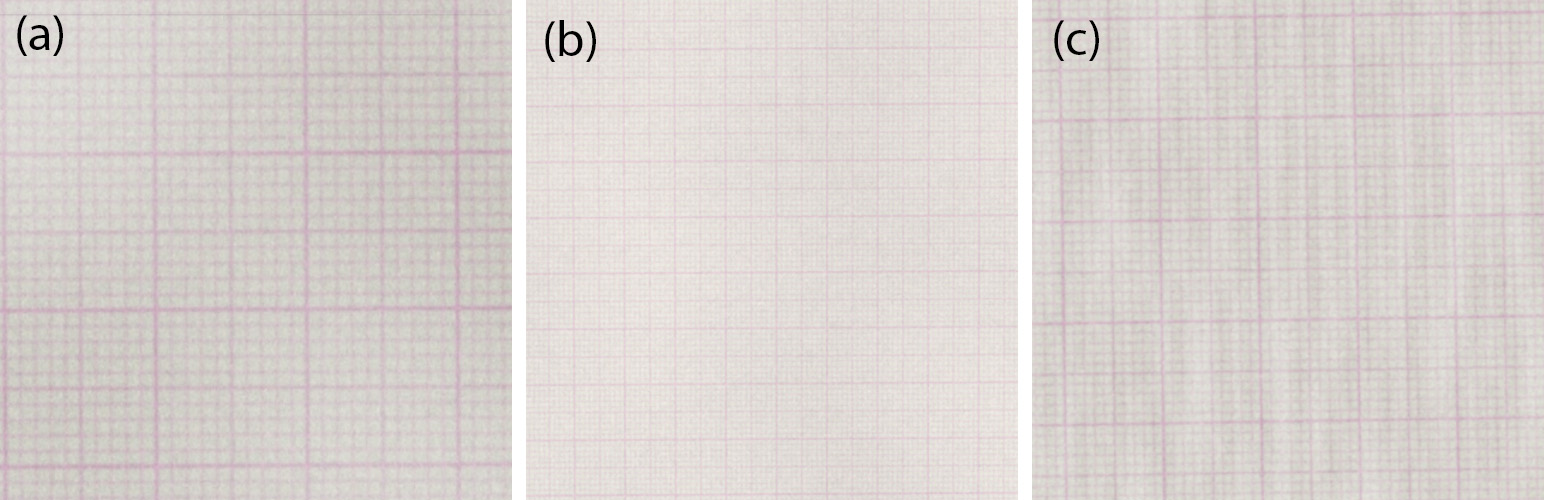}}
\caption{Selection of different millimeter papers generated using our procedural shader method ranging from sharp to blurry.}
\label{fig:mm_paper}
\end{figure}

\subsection{Procedural Millimeter Paper Model}
The procedural millimeter paper model generates unique millimeter paper textures, serving as backgrounds in scenes for renderings of leaves.

Our method is implemented in the fragment shader by procedurally modeling a texture of standard millimeter paper using sine functions. Given a local space position of a fragment \((x, y)\) on the millimeter paper, the color intensity is determined by the following function:
\begin{equation}
    C(x, y, \phi) = A \cdot \sin(B \cdot x + \phi) + D\,,
\end{equation}
where \( A \) is the amplitude of the stripes, which controls their intensity variation, \( B \) is their frequency, which determines the distance between them, \( \phi \) is the phase shift which offsets them horizontally, and \( D \) is the baseline color intensity.

Further shader effects include hue, contrast, brightness, and saturation changes to emulate various paper conditions. Subsequently, we blend this texture with multiple layers of noise. We use three types of noise: gradient \( G \), Voronoi \( V \), and simple \( S \). A selection of rendering results created by different parameter value configurations is shown in Fig.~\ref{fig:mm_paper}. 

% \textit{Gradient Noise} is described as 
% \begin{equation}
%   G(x, y) = \nabla P(x, y)\,,
% \end{equation}
% where $x,y$ is the 2D location. It is applied with varying strength and adds subtle variations to mimic paper inconsistencies.

% \textit{Voronoi Noise} is described by the following equation:
% \begin{equation}
%   V(x, y, \theta) = \sum_{i,j} f(x - x_{i,j}, y - y_{i,j}, \theta)\,,\label{eqn:vor}
% \end{equation}
% where $\theta$ is the angle. It is characterized by random angles and densities, and it contributes to texture complexity.

% \textit{Simple Noise} is described by: 
% \begin{equation}
%   S(x, y) = N\left(\frac{x}{w}, \frac{y}{h}\right)\,
% \end{equation}
% where $w$ and $h$ are the width and height parameters.
% The simple noise is defined by random width and height and then rescaled, introducing more pronounced textural changes.

The composition of these noise layers is expressed as a weighted sum:
$$
L(x,y)=w_G\cdot  G(x,y)+w_V\cdot  V(x,y)+w_S\cdot  S(x,y)\,,
$$
where $w$ are the weights determining the strength of each noise.

%Figure \ref{fig:mm_paper} showcases the diversity of our procedural model in generating millimeter paper textures, labeled as (a), (b), (c), and (d), each reflecting different noise combinations. Case (a) employs minimal gradient noise for a pristine, new paper look. Case (b) combines moderate gradient and Voronoi noises, suggesting slightly aged or handled paper with subtle imperfections. In case (c), increased simple noise intensity alongside gradient and Voronoi noises creates a texture of more extensively used paper, with characteristics like minor creases. Case (d) represents heavily worn paper, achieved through high levels of all three noise types, resulting in a distinctly weathered and textured appearance. These variations in Figure \ref{fig:mm_paper} illustrate our model's ability to replicate a spectrum of real-world paper conditions.

\subsection{Procedural Leaf Model}
Our algorithm for generating leaf models procedurally incorporates several computation stages to simulate leaf morphology. The initial shape is defined on the CPU using Unity's animation curve, which is a parametric piecewise polynomial curve defined by a set of control points that interpolate values to form a smooth transition. 

Random perturbations are applied to the control points' positions to introduce variations, reflecting the inherent diversity found in leaf shapes. This randomness is mathematically expressed by adding a noise function \( N \) to the control point positions \( P \), where the new position \( P' = P + N \).

\begin{figure}[hbt]
\centerline{\includegraphics[width=0.99\linewidth]{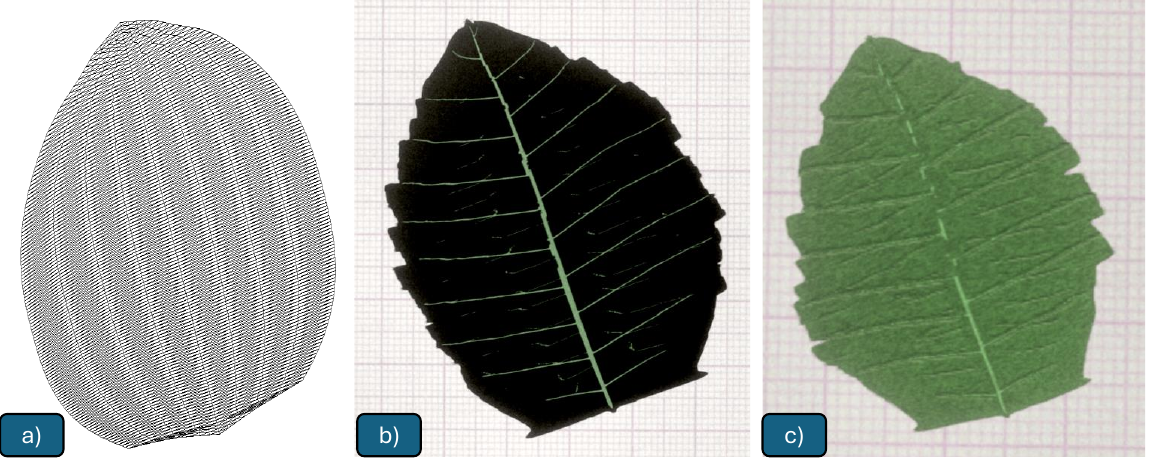}}
\caption{Procedural leaf model generation: The shape is defined by a parametric curve using Unity's Animation Curve (a), which is then textured, including vein pattern development (b), and stochastic elements and surface details are added via shader effects (c).}
\label{fig:leaf_model}
\end{figure}

For vein patterning, the algorithm follows the method outlined by \citeauthor{goldman2004turtle} \cite{goldman2004turtle}, employing a turtle graphics system. This system is formalized as a series of commands defining the turtle's movement, where the turtle's path through space traces the veins - generated in screen space as a procedural texture. We add a stochastic element by incorporating Brownian Motion \( Bm(t) \) for jittering the turtle's movement to create wavy vein lines and randomness in the branching angles~\( \theta_{veins} \) for the vein paths.

Texture generation extends to creating a height map~\( H \) for depth representation, where \( H \) is modified by a function of the turtle's path and its corresponding vein thickness. The height map is used to obtain normals for the normal mapping within the fragment shader to produce detailed leaf surface textures (examples shown in Fig.~\ref{fig:leaf_model}).

Vertex displacements in the mesh are introduced using the Voronoi noise function \( V \)  within the vertex shader to simulate the leaf's surface undulations. The fragment shader then employs a blend of procedural noises and a photorealistic leaf texture. Initially, we define two primary color textures, \( C_1 \) and \( C_2 \), representing different aspects of leaf coloration and patterning. The final leaf texture \( T \) is the result of blending these color textures, modulated by gradient noise \( G(x, y) \). Additionally, detailed features are incorporated: veins (denoted as \( V_e \), holes (denoted as \( H_o \)), and other textural elements such as spots and edge irregularities, which mimic natural imperfections in leaf morphology.

\begin{figure}[hbt]
\centering
\includegraphics[width=\linewidth]{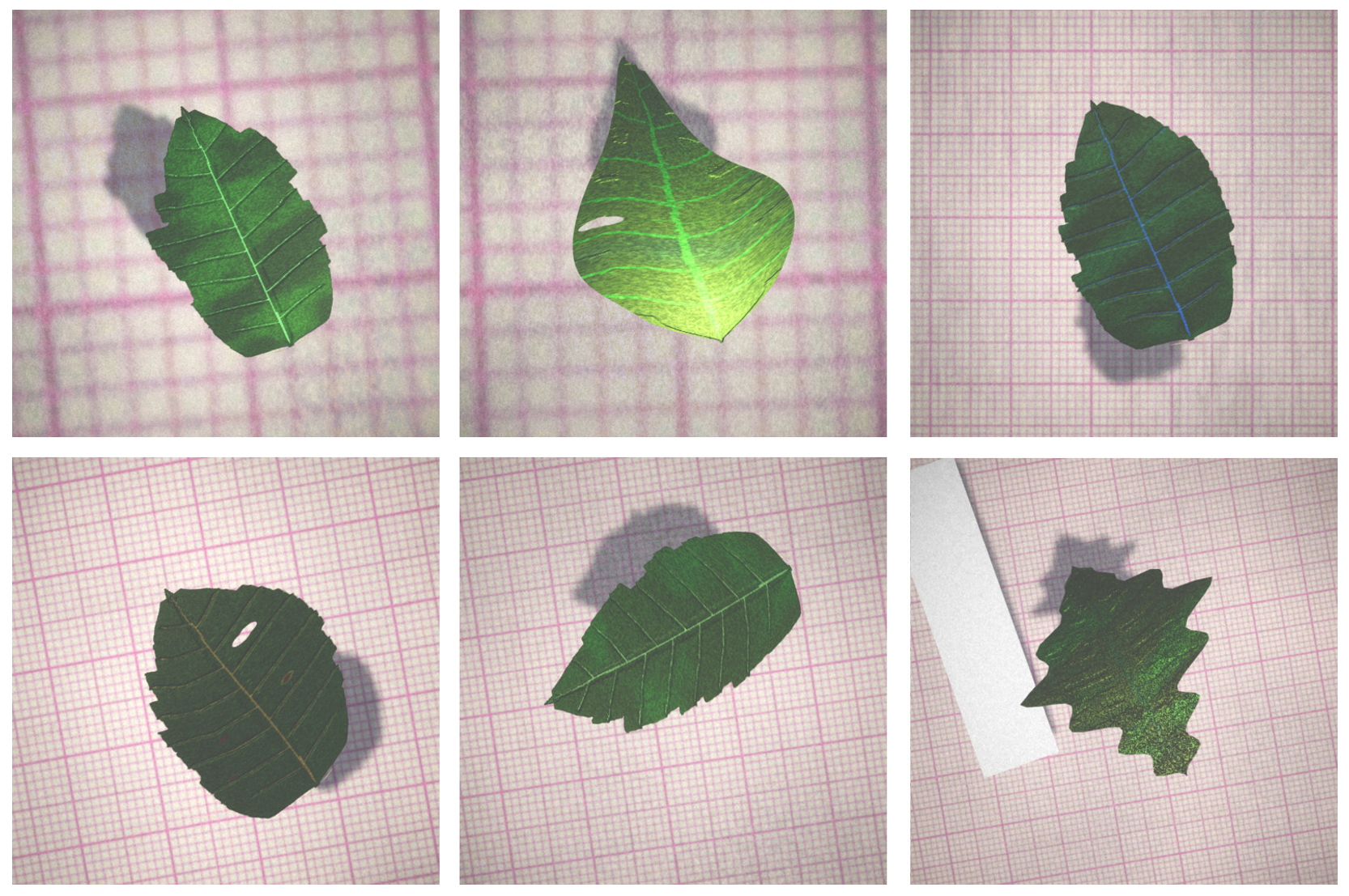}
\caption{Diverse final renderings from the procedural leaf generation pipeline. This collection illustrates the variation achieved in leaf appearance through our procedural model parameters. Each rendering captures different lighting conditions, shadow effects, and background scaling.}
\label{fig:final_renderings}
\end{figure}

\section{Implementation}

\paragraph{Rendering} Each leaf undergoes four separate rendering passes to capture varied appearances (examples shown in Fig.~\ref{fig:final_renderings}). In each pass, we adjust shader parameters for shadow rendering, including strength (\(s\)), position (\(\vec{p}\)), and size (\(sz\), referring to shadow size), to simulate various lighting environments. These parameters are governed by a shadow function \(S(s, \vec{p}, sz)\), which is implemented using the shadow mapping algorithm. 

\paragraph{Background Scaling}
The size of the millimeter paper in the scene is adjusted via a scaling factor \( \gamma \), maintaining a consistent camera perspective across all renderings. The scaled background is denoted as \( M' = \gamma M \).

\paragraph{Scene Composition}
Additional elements, like paper fragments and glass, are included to enhance realism. Their transformations in the scene are handled through a combination of randomized translations (\( T_{xy} \)), rotations (\( R_{\theta} \)), and scalings (\( S_{xy} \)).

%\paragraph{Data Normalization}
%To address data distribution anomalies in the size attribute, we apply a root transformation (\( \sigma' = \sqrt{\sigma} \)). This normalizes the distribution, ensuring better correlation with empirical data and improved model training and evaluation reliability. \mc{What is this doing exactly? Is \(\sigma'\) the standard deviation of some parameters?}

\subsection{Data Preparation}
An important aspect of our dataset is the generation of semantic masks and the precise computation of leaf area size labels. These elements are essential for various applications, including detailed morphological analysis and machine learning model training.

We initiate another pass of the graphics pipeline for each leaf rendering with a uni-colored leaf mesh on a black background to obtain a semantic mask. The computation of the leaf area size is a direct application of the procedural parameters used in our millimeter paper model. Given that the paper model's grid is generated with known dimensions, and the scaling factor (\(\gamma\)) used in rendering is also known, we can accurately compute the area of each leaf by summing the areas of all triangles comprising the leaf's surface mesh. 

%BB - I think we do not need to explain how to sum triangles...
%Specifically, the leaf area \( A \) is calculated as :
%\begin{equation}
%A = \sum_{i=1}^{n} A_i
%\end{equation}
%where \( A_i \) is the area of the \( i \)-th triangle in the mesh, and \( n \) is the total number of triangles. The area of each triangle is computed using the standard formula for the area of a triangle in 3D space, which can be readily calculated given the vertices of each triangle.

% \begin{figure}[htbp]
% \centering
% \includegraphics[width=\linewidth]{figs/}
% \caption{Leaf Mask.}
% \label{fig:combined_rendering}
% \end{figure}

% \begin{figure}[t]
% \centering
% \includegraphics[width=\linewidth]{combined_rendering.jpg}
% \caption{The left side depicts the detailed rendering of the leaf, while the right side shows the pixel-precise semantic mask.}
% \label{fig:combined_rendering}
% \end{figure}

\subsection{Integration of ControlNet Inpainting}
Following the rendering of an initial 100,000 annotated synthetic images along with their masks, we use these images to obtain realistic 3D annotated counterparts. To achieve this objective, a pre-trained ControlNet network by \citeauthor{zhang2023controlnet}~\cite{zhang2023controlnet} is utilized. 

In our implementation, the Stable Diffusion model \cite{rombach2022high} serves as the backbone, characterized by a U-Net structure comprising an encoder, a middle block, and a skip-connected decoder.

The model employs a combination of down-sampling and up-sampling convolutions, ResNet blocks, and Vision Transformer layers for feature extraction and manipulation. Textual inputs are encoded using a CLIP model, which is necessary for the generation process toward specific text-described attributes \cite{radford2021learning}.

\begin{figure}[t]
\centering
\includegraphics[width=\linewidth]{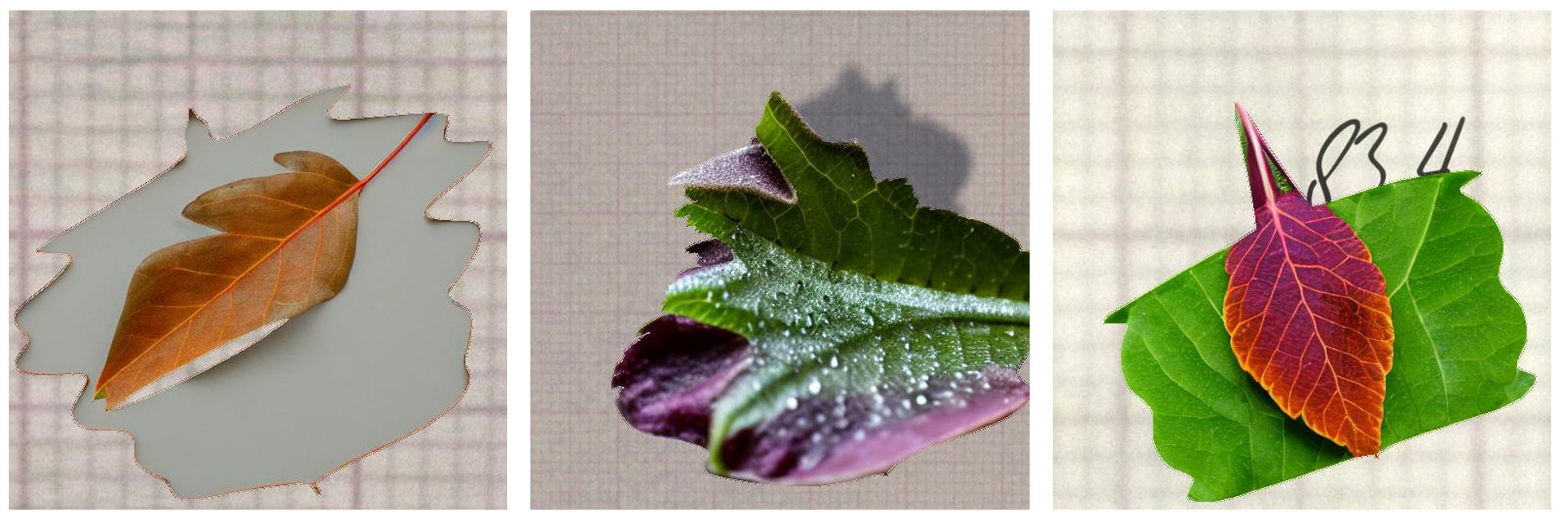}
\caption{Three instances of ControlNet-generated images where the region of the inpainted leaf in the mask deviates significantly from the region defined by the procedurally generated mask. Such data points are automatically filtered out in LAESI.}
\label{fig:Faulty_Images}
\end{figure}

ControlNet provides a range of trained networks with diverse image-based conditions to regulate the diffusion models. These conditions include edges generated by various methods, depth, and normal maps, human poses, semantic segmentation, user sketches, etc. In our experiments, we employed the Canny-edge detection method \cite{canny1986computational}. Specifically, to generate realistic images from each synthetic image, we create an image using the Canny edges of our synthetic masks as inputs. The text prompt input consists of the specific phrases ``oak leaf on millimeter paper'' and ``beech leaf on millimeter paper'', which provide input to ControlNet to inpaint the desired attributes of the synthetic leaf images. By incorporating this textual information, ControlNet can tailor the inpainting process to match the characteristics of oak or beech leaves. We then replace the background with our procedurally generated background. Note that we cannot use the millimeter paper background from the AI-generated image, or the leaf surface area annotations would become inconsistent. 

\begin{figure}[hbt]
  \centering
  \includegraphics[width=\linewidth]{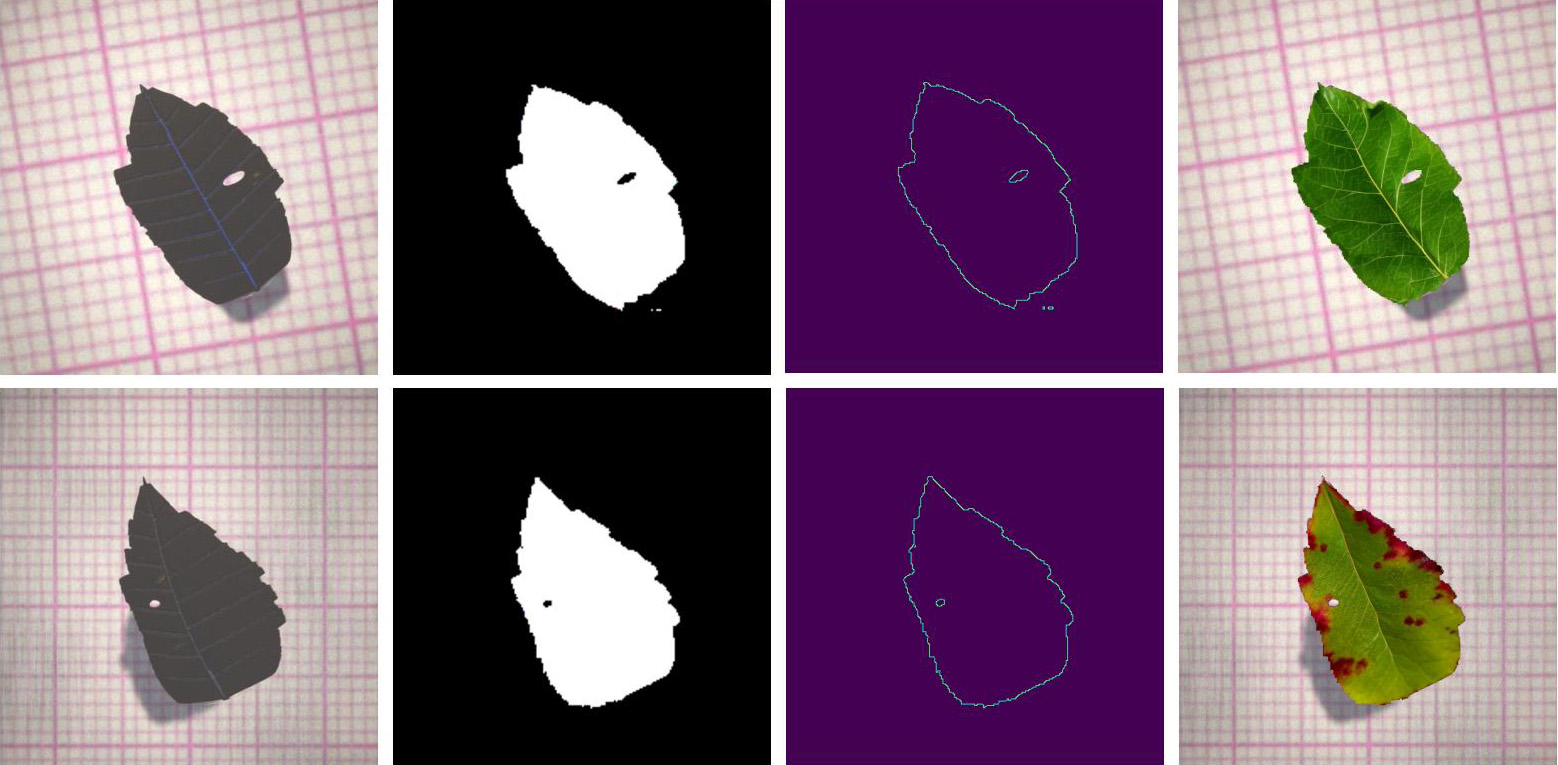}
  \caption{Example of inpainted results using ControlNet for semantic mask inpainting. In the lower row, the inpainting resulted with the addition of disease features, which were not described with the procedural model and would have been very challenging to simulate procedurally.}
  \label{fig:inpainted_results}
\end{figure}

Post-generation with ControlNet, the synthetic data undergoes a filtering process to eliminate images inconsistent with the associated annotations. This dataset refinement utilizes a MobileNet-based semantic segmentation model, trained on the 'synthetic rendering 2' dataset, comprising 5,000 synthetic and 1,700 real images (see Tab.\ref{tab:results}). Synthetic images, where the predicted mask deviates by more than 15\% from the ground truth, such as in instances shown in Fig.~\ref{fig:Faulty_Images}, are filtered out of the dataset. This step removes outliers that we encountered during the ControlNet inpainting step in the frequency of 15-20\%.

\begin{figure}[t]
  \centering
  \includegraphics[width=\linewidth]{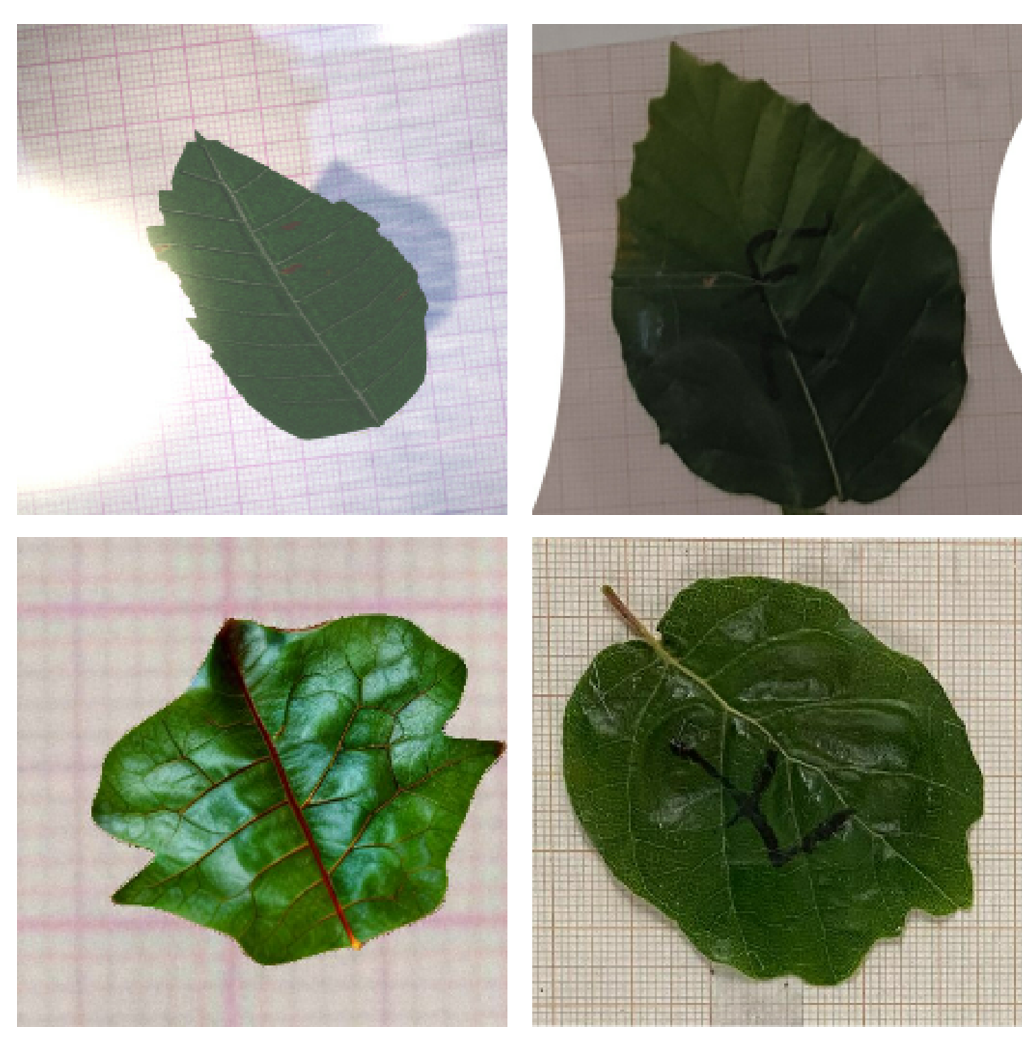}
  \caption{Two pairs of synthetic (left) and real (right) images selected from the 100 highest cosine similarity scores from the Rendering 2 dataset and below from the ControlNet+Filtering dataset. }
  \label{fig:comparison_closest}
\end{figure}

\begin{figure}[htbp]
\centering
\includegraphics[width=\linewidth, height=\linewidth]{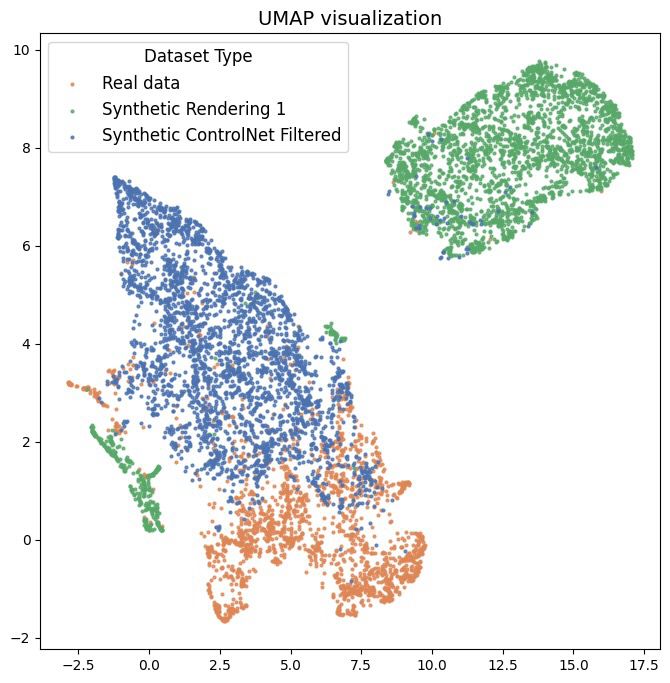}
\caption{UMAP visualization of the ResNet50 (CLIP ViT-B/32) embeddings of real data (orange) and two different synthetic image sets (Rendering 1 - green, ControlNet + Filtering - blue). The lack of separation in the feature space between blue and orange dots suggests that the synthetic images for the ControlNet + Filtering dataset contain semantically more similar features compared to the Rendering 1 dataset.}
\label{fig:cosine_similarity}
\end{figure}

\begin{figure}[htbp]
\centering
\includegraphics[width=\linewidth, height=0.5\linewidth]{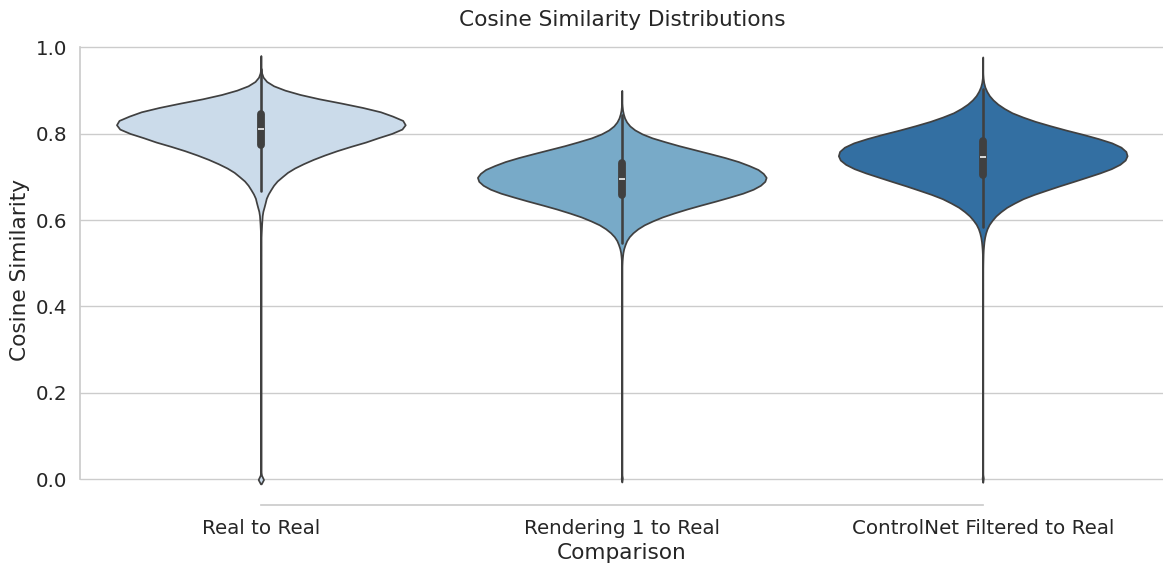}
\caption{Violin plots of cosine similarity scores for datasets used in Fig.~\ref{fig:cosine_similarity}. ControlNet Filtered image distribution has overall higher cosine similarity scores compared to the Rendering 1 dataset. The distributions are obtained from images which have overall similar features compared to real ones as indicated by high scores.}
\label{fig:cosine_similarity2}
\end{figure}

\section{Validation}\label{sec:validation}
Here, we present the validation of synthetic data for training deep learning models. Specifically, we apply synthetic data to train leaf area size prediction and semantic segmentation of real leaves on mm paper. This vision task is made deceptively difficult by using different sensors, camera extrinsic, and other image artifacts, such as reflections, notes, and objects appearing in the photographs in a real research environment. The application of rule-based leaf area prediction methods usually relies on either highly controlled environments~\cite{RAHIMIKHOOB2023113636} or specific reference objects included for scaling purposes~\cite{easlon2014easy}, which makes these methods impractical for typical research compilation efforts.

\subsection{Network Model Training}
We used the MobileNet V3 architecture~\cite{howard2019mobilenetv3} to predict leaf area size through regression. The choice of MobileNet V3, specifically designed for mobile applications running on hand-held devices such as smartphones, aligns with the goal of deploying our solution in remote locations. We implemented our framework in Python, adopting TensorFlow libraries for deep learning.

\subsection{Hyperparameter Optimization}
Several hyperparameters were systematically varied during the experiments, including (1) Architecture variations (MobileNet V3 Large, MobileNet V3 Small). (2) Data augmentation techniques (brightness, contrast, hue, saturation, flip, 90$^o$ rotation, random rotation, Gaussian and Poisson noise). Due to lack of identification of optimal hyperparameter configuration in the scientific literature, we employed a large hyperparameter space exploration by conducting 1,425 experiments to identify the most performant models through crowd sourcing (matrix of hyperparameter value configurations included in supplementary material).

Optimal hyperparameters were identified based on validation dataset performance. The best settings were MobileNet V3 Large with ImageNet transfer learning (pre-trained with ImageNet-1k weights), RMSprop optimizer, an initial learning rate of $1e^{-3}$ %0.001 
with Piecewise Constant Decay down to $1e^{-9}$ %0.00000001, 
and augmentations including brightness, contrast, hue, saturation, flip, 90$^o$ rotation, and Poisson noise.

\begin{figure}[t]
\centering
\includegraphics[width=\linewidth]{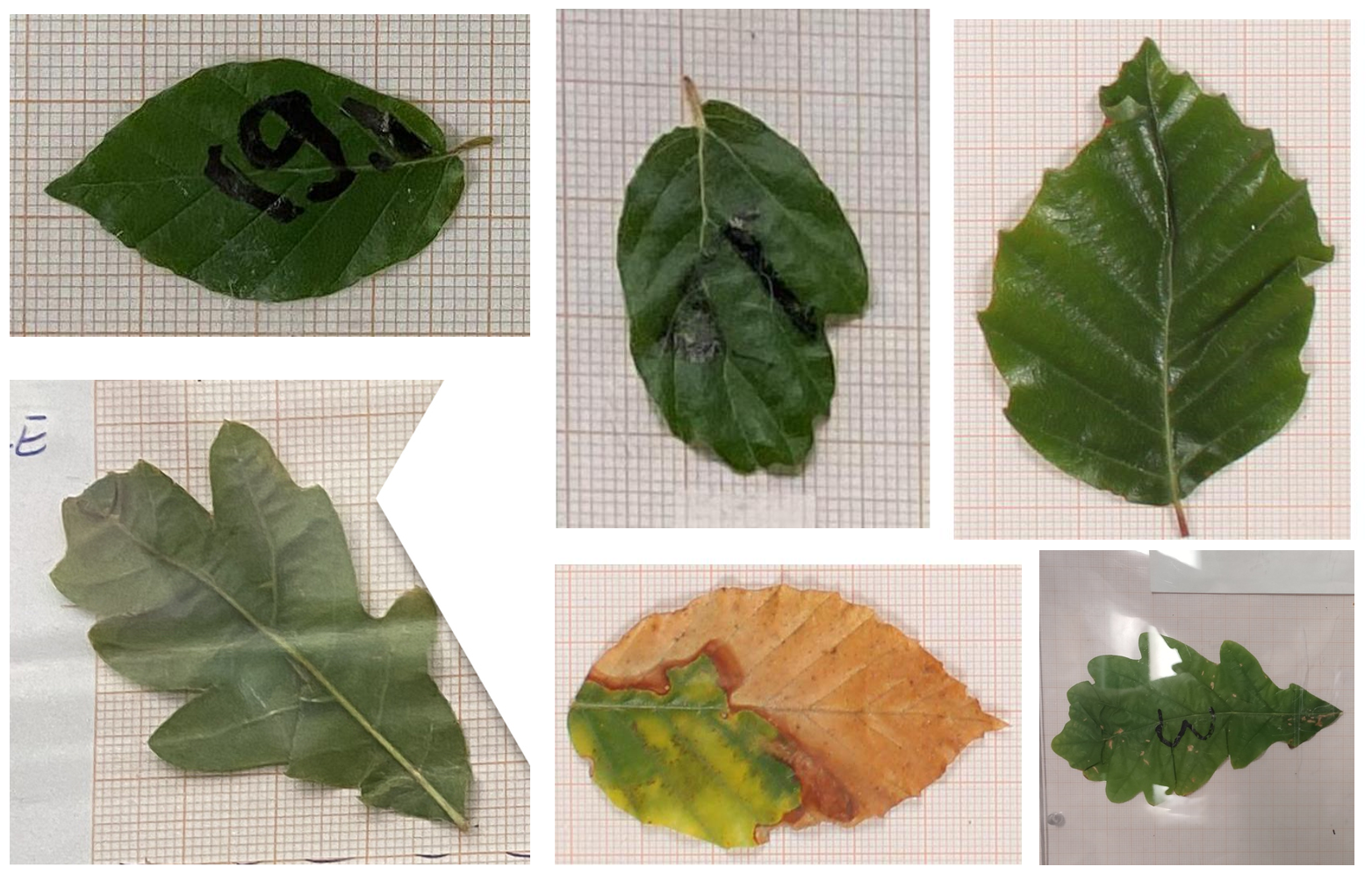}
\caption{Photographs of beech and oak leaves on millimeter paper taken at the TUM School of Life Sciences. They were used to create a baseline for training a leaf surface area prediction and semantic segmentation network models and for the validation dataset.}
\label{fig:leaves_on_mm_paper}
\end{figure}

\subsection{Validation Experiments}
We conducted six training experiments:
\begin{enumerate}
  \item \textit{Rule-based Baseline}: 30 real data points with uniformly-sized red squares for calibration purposes in the images. We calculated performance metrics with the "Easy Leaf Area" rule-based leaf area prediction method ~\cite{easlon2014easy} in contrast to all other experiments which use the MobileNet V3 model.
  \item \textit{Real Data Baseline}: Training with 1,7K real, annotated leaf data points (examples shown in Fig.~\ref{fig:leaves_on_mm_paper}). 
  \item \textit{Synthetic Rendering 1}: Training combined 1,7K real data points with 5K synthetic data points. 
  \item \textit{Synthetic Rendering 2}: Training combined 1,7K real data points with 5K synthetic data points using improved parameter value configurations based on results obtained with previous dataset. Specifically, we changed model parameters for the millimeter paper model remove very noisy paper effects by narrowing parameter value ranges of noise functions.
  \item \textit{Synthetic ControlNet}: Training combined 1,700 real data points with 5K, 6K, 7K, 8K, 9K, and 10K synthetic data points, using ControlNet for mask inpainting.
  \item \textit{Synthetic ControlNet + Filtering}: Training used ControlNet and a filtering process 5K, 6K, 7K, 8K, 9K, and 10K synthetic data points, mixed with 1,7K real data points.
\end{enumerate}

Each experiment utilized the same validation and test datasets comprising 250 real annotated photographs each. The validation data has been harvested from climate chamber experiments conducted at the TUMmesa ecotron and at the Technical University in Munich (TUM) Plant Technology Center in the years 2021 and 2022. Real annotations were obtained empirically and with the LICOR LI-3100C Area Meter.

\subsection{Evaluation Metrics}
The primary metric for evaluation is the mean relative error (MRE) of the leaf area size between predicted and ground truth values. For the semantic segmentation metric we employed the mean intersection over union (mIoU) of the ground truth to predicted masks, as well as the relative error in total count of mask pixels - called mask pixel error (MPE). Furthermore, we employ cosine similarity scores and the UMAP \cite{mcinnes2018umap} dimensionality reduction method to quantify the similarity of synthetic and real images (Figs.~\ref{fig:comparison_closest}-~\ref{fig:cosine_similarity2}).

\begin{figure}[t]
\centering
\includegraphics[width=\linewidth]{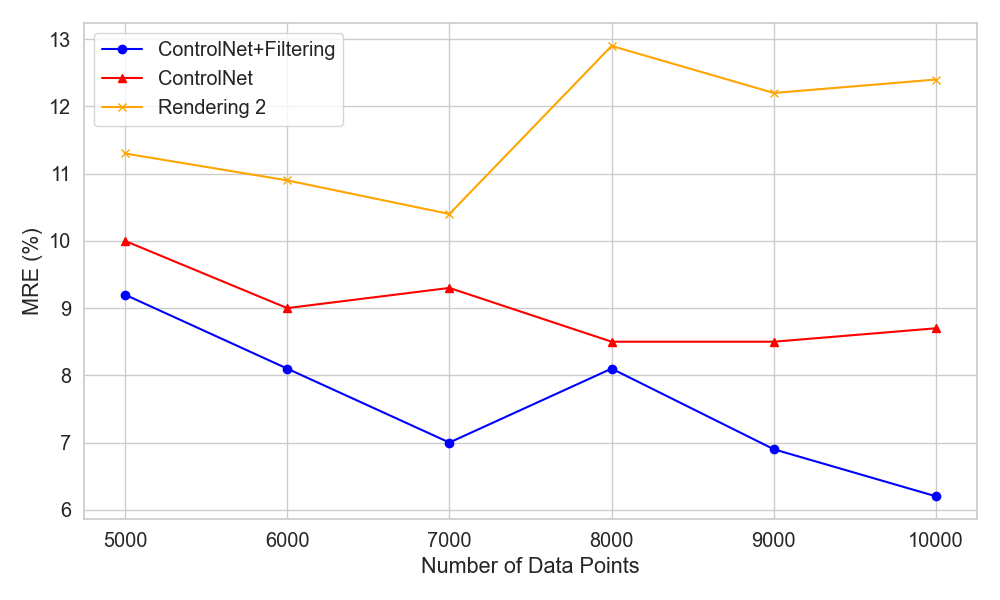}
\caption{Validation data loss curves for experiments for training with datasets ranging from 5K to 10K data points. Red curve indicates results obtained with raw ControlNet inpainted data. blue curve with filtered data, and orange without inpainting (Rendering 2). While the addition of more data significantly improves the MRE on leaf area size prediction on inpainted data, there is no improvement for raw synthetic data.}
\label{fig:training_validation_loss}
\end{figure}

The outcomes of our synthetic training experiments, summarized in Table~\ref{tab:results}, reveal significant differences in model performance based on the type and amount of synthetic data used in training network models. The Real Data Experiment, serving as a baseline, achieved an MRE of above 12.5\% in validation and test sets. Notably, Rendering 1+2 experiments with mixed-in synthetic data demonstrated a slight improvement in validation MRE. However, the most significant advancements were observed in the Synthetic ControlNet and Synthetic ControlNet + Filtering experiments. Here, the inclusion of 10,000 synthetic data points, coupled with filtering, substantially reduced the MRE to as low as 6.1\% in validation and 6.2\% in tests (Fig.~\ref{fig:training_validation_loss}). Interestingly, increasing the amount of synthetic data results in a proportionally greater improvement in MRE for the experiment with filtering compared to the other experiments. Furthermore, while the MRE for leaf area prediction improved for inpainted data (with and without filtering) the performance as measured by mIoU and MPE for the semantic segmentation model decreased for the ControlNet experiment (0.09 MPE) compared to the non-inpainted experiments (0.07 MPE) but increased for the ControlNet+Filtering experiment (0.05 MPE, Tab.~\ref{tab:results}). This shows the usefulness of filtering after the inpainting step in LAESI as it significantly improves performance across the two visual downstream tasks.

\begin{table*}[t]
\centering
\caption{Performance comparison of training experiments with real baseline (1,7K) and 5K synthetic training data, and a rule-based baseline.}
\label{tab:results}
\begin{tabular}{lcccc}
\toprule
\textbf{Experiment} & \textbf{Validation MRE} (\%) & \textbf{Test MRE} (\%) & \textbf{mIoU} (\%) & \textbf{Mask Pixel Error} (\%) \\
\midrule
Rule-based Baseline & 38.3 & 38.3 & - & - \\
Real Data Baseline & 12.5 & 12.9 & 0.79 & 0.08 \\
Rendering 1 & 12.0 & 11.0 & 0.81 & 0.07 \\
Rendering 2 & 10.8 & 11.3 & 0.82 & 0.07 \\
ControlNet & 8.4 & 10.0 & 0.8 & 0.09 \\
ControlNet + Filtering & 8.5 & 9.2 & 0.83 & 0.05 \\
\bottomrule
\end{tabular}
\vspace{-2mm}
\end{table*}

\section{Discussion and Conclusion}\label{sec:conclusion}
This study presents a fully automatic approach for generating synthetic data for leaf area prediction by utilizing procedural models and the MobileNet V3 neural network architecture. While standard computer vision methods for leaf area prediction (e.g., \cite{easlon2014easy}) can work well in highly controlled contexts, we found that typical photographs compiled from wet lab experiments contain unexpected complexity making such methods challenging to use in practice. This includes notes, reflections, variable sensor intrinsics and extrinsics, pieces of paper, and other objects present in the images (see Fig.~\ref{fig:inpainted_results}). Further, our results indicate that the inpainting of synthetic data with ControlNet is a feasible way for improving performance on a specialized vision task, and that the addition of filtering into the generative pipeline based on annotation consistency significantly improves overall performance. Specifically, our results indicate that the synthetic data before inpainting likely lacks sufficient feature variability to benefit training models on increasing quantities of datapoints (Fig.~\ref{fig:training_validation_loss}, orange curve), while the non-filtered inpainted data decreases model performance in the semantic segmentation task by introducing inconsistencies to annotations (Tab.~\ref{tab:results}, mIoU and MPE scores). Surprisingly, this means that the training on erroneous annotations in the ControlNet experiment still leads to better overall MRE in leaf area size prediction. Overall, our analyses indicate that the best domain adaptation for our procedurally generated synthetic data can be achieved via filtered inpainting which reduces annotation inconsistencies while preserving the AI-generated features. These findings are similar to those of Fei et al. \cite{Fei2021synthetic2real} who show that using a semantic constraint loss in the training of GAN-based methods can help maintaining annotation consistency for improved model performance.

Through a series of experiments, we evaluated various hyperparameter value configurations and datasets, leading to an optimal setup that achieved a test loss of 6.2\% in our final experiment, which is at least on par with the human annotation error and significantly outperforms the best real baseline. This led to the adoption of this model by TUM biology researchers in further experiments, making manual annotation redundant. The manual annotation of over 2,000 leaf images is a costly and laborious undertaking, prolonging the compilation of empirical results. Our results prove the value of synthetic data in training deep learning models for a specific research application, strengthening the claims of the usefulness of synthetic data made in other research work (e.g.,  \cite{fan2023scaling,shipard2023diversity,yu2023diversify, klein2023synthetic}
).

\begin{figure*}[htbp]
\centering
\includegraphics[width=\linewidth]{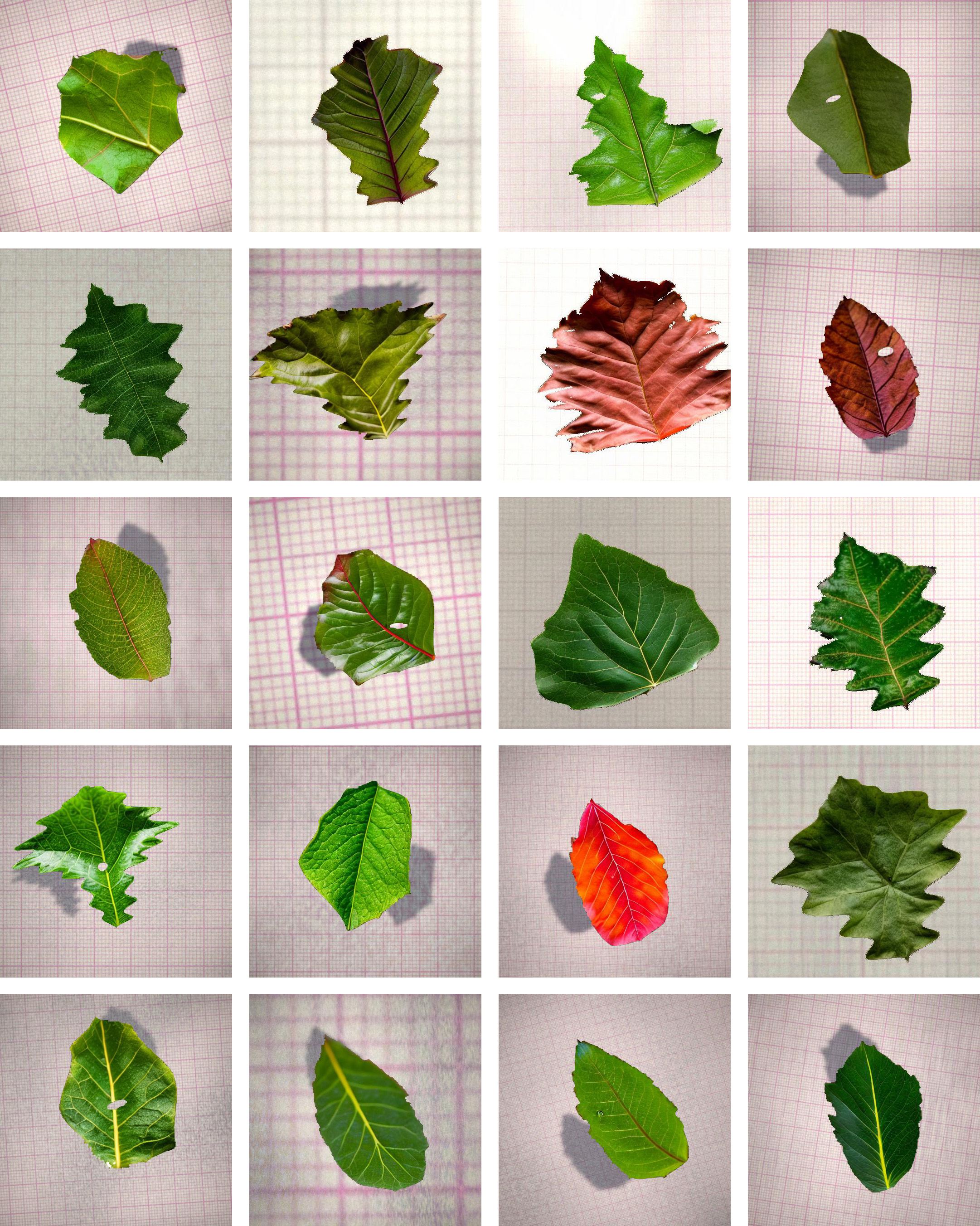}
\caption{A selection of synthetic images generated with LAESI. These images are part of the ControlNet + Filtering subset.}
\label{fig:results}
\end{figure*}

Furthermore, our approach, which combines procedural generation with generative AI methods, holds promise for various other applications in botanical research and agriculture, particularly in remote sensing and precision agriculture. As future work, we would like to address the estimation of other growth parameters, such as shoot internode length or root biomass. Our approach can be extended to other domains, allowing fully automatic synthetic data generation for machine learning.

%Future work will focus on refining the models further, exploring additional plant species, and integrating the developed system into mobile platforms for real-time field applications. The success of this research opens avenues for more extensive adoption of synthetic data in machine learning, enabling innovative solutions in environmental and agricultural technologies.

{\small
\bibliographystyle{ieeenat_fullname}
\bibliography{11_references}
}

\ifarxiv \clearpage \appendix %\section{Appendix Section}
%\label{sec:appendix_section}
%Supplementary material goes here.
 \fi

\end{document}

%% TITLE
\title{\paperTitle}
\author{\authorBlock}
\maketitlesupplementary

\appendix
%\section{Appendix Section}
%\label{sec:appendix_section}
%Supplementary material goes here.

{\small
\bibliographystyle{ieeenat_fullname}
\bibliography{11_references}
}